\documentclass{article}

\usepackage{arxiv}

\usepackage[utf8]{inputenc} 
\usepackage[T1]{fontenc}    
\usepackage{hyperref}       
\usepackage{url}            
\usepackage{booktabs}       
\usepackage{amsfonts}       
\usepackage{nicefrac}       
\usepackage{microtype}      
\usepackage{lipsum}
\usepackage{graphicx}
\graphicspath{ {./images/} }
\usepackage{amsmath}
\usepackage{amssymb}
\usepackage{booktabs}
\usepackage{url}
\usepackage{hyperref}
\usepackage{listings}
\usepackage{xcolor}
\usepackage{algorithmic}
\usepackage{algorithm}
\usepackage{comment}
\usepackage{multirow}
\usepackage{comment}

\hypersetup{
    colorlinks=true,
    linkcolor=blue,
    citecolor=blue,
    urlcolor=blue
}
\usepackage[most]{tcolorbox}

\definecolor{lightyellow}{RGB}{255, 255, 204}

\usepackage[table]{xcolor}
\usepackage{array}
 
\definecolor{headergray}{gray}{0.97}

\title{Why We Need World Models for AGI: Where LLMs Fail and How World Models May Outperform}

\author{
Feisal Alaswad \\
Department of Computing Technologies\\
SRM Institute of Science and Technology\\
Kattankulathur 603203, Tamil Nadu, India \\
\texttt{feisal.alaswad@hotmail.com} \\
\And
Batoul Aljaddouh \\
Department of Computing Technologies\\
SRM Institute of Science and Technology\\
Kattankulathur 603203, Tamil Nadu, India \\
\texttt{batoul.aljaddouh@gmail.com} \\
\And
Maher Alrahhal \\
Bio-Sensing and Bio-Sensors Group\\
Smart Automation and Communication Technologies\\
Research Institute of Sciences and Engineering\\
University of Sharjah, UAE \\
\texttt{malrahhal@sharjah.ac.ae} \\
\And
Poovammal E \\
Department of Computing Technologies\\
SRM Institute of Science and Technology\\
Kattankulathur 603203, Tamil Nadu, India \\
\texttt{poovamme@srmist.edu.in} \\
\And
Talal Bonny \\
Department of Computer Engineering\\
College of Computing and Informatics\\
University of Sharjah, UAE \\
\texttt{tbonny@sharjah.ac.ae} \\
}

\begin{document}
\maketitle

\begin{abstract}
Large language models achieve strong performance in language generation and knowledge-intensive tasks, yet remain limited in settings requiring causal reasoning, persistent state tracking, and long-horizon planning. We argue that these limitations may arise from an objective-level mismatch between sequence prediction and reasoning over latent environment dynamics. To formalize this distinction, we introduce Latent Dynamics Inference (LDI), a conceptual perspective that interprets language and multimodal observations as partial evidence of underlying transition dynamics.

To empirically investigate this perspective, we introduce \textsc{Flux}, a sequential reasoning environment specified entirely through natural-language rules. As a proof-of-concept case study, the rules are first compiled into an explicit state-transition simulator, illustrating that structured latent transition dynamics can, in some cases, be operationally extracted from textual rule descriptions. This enables a controlled comparison between the LLMs operating purely over textual observations and reinforcement-learning agents trained directly within the extracted latent state space. Within this case study, agents operating with explicit access to the latent state space exhibit substantially more stable behavior in long-horizon gameplay, achieving an aggregate win rate of approximately 79\% versus 11\% for LLMs. Qualitative analysis further reveals failure modes consistent with unstable persistent state tracking, including invalid actions, state-tracking errors, and short-horizon reasoning failures.  The complete implementation of the \textsc{Flux} environment available at \href{https://github.com/FeisalAlaswad/FLUX-RL-Agent}{\texttt{https://github.com/FeisalAlaswad/FLUX-RL-Agent}}.

Within the evaluated setting, these results suggest that strong sequence prediction alone may struggle to support robust long-horizon dynamic reasoning without mechanisms for persistent state tracking and transition modeling.
\end{abstract}


\section{Introduction}

Models trained with large-scale autoregressive learning objectives have demonstrated remarkable performance across a wide spectrum of tasks, including natural language understanding, code generation, and multi-step reasoning. These models leverage vast textual corpora to learn rich statistical representations of language, enabling impressive generalization within the distribution of observed data~\cite{lukasik2025better,shi2025continual,zhang2024out,kumar2024large,aljaddouh2026multimodal}.

Despite these advances, a growing body of evidence indicates that LLMs exhibit systematic limitations when deployed in settings requiring deeper forms of intelligence~\cite{shojaee2025illusion,wong2025reasoning,momennejad2023evaluating,alaswad2026cocomo,gendron2023large}. In particular, their performance degrades in scenarios involving:

\begin{itemize}
    \item \textbf{Causal reasoning and counterfactual inference}, where understanding the effect of interventions is required~\cite{kasetty2024evaluating,bhasuran2026evaluation,chen2025counterbench,ashwani2024cause}.
    \item \textbf{Long-horizon planning} in complex and partially observable environment~\cite{sinha2025illusion,vinay2025failure,wu2024mldt}.
    \item \textbf{Grounded interaction} with physical or simulated agents~\cite{carta2023grounding,xiang2023language}.
    \item \textbf{Environment simulation} and predictive modeling of future states~\cite{carta2023grounding,farkavs2025will}.
    \item \textbf{Stateful decision-making} under uncertainty and feedback~\cite{yang2025feedback}.
\end{itemize}

These limitations suggest an important gap: high-quality sequence prediction does not necessarily translate into robust modeling of how systems evolve over time. In many real-world settings, effective reasoning requires modeling state transitions, causal structure, and the consequences of actions within dynamic environments~\cite{ge2026review}.

This motivates the study of world models: systems that maintain explicit latent state representations together with transition dynamics capable of modeling state evolution under actions or interventions~\cite{petri2025learning,ha2018world}. Unlike autoregressive sequence models, world models operate over latent environment states rather than observation sequences alone, making causal prediction, simulation, and planning structurally more natural.

More broadly, these limitations raise fundamental questions about the suitability of autoregressive sequence modeling as a foundation for Artificial General Intelligence (AGI). Many formulations of general intelligence emphasize not only linguistic fluency, but also causal reasoning, planning, and environment interaction, long-horizon planning, and the ability to maintain coherent internal models of evolving worlds. Importantly, a competing view holds that sufficiently scaled LLMs may approximate world-model-like behavior through emergent capabilities. 

However, we argue that these limitations may partly arise from a mismatch between objectives optimized for sequence prediction and those required for explicit environment modeling. Although sufficiently scaled language models may exhibit behaviors resembling planning or causal reasoning in constrained settings, such behavior is not explicitly enforced by the autoregressive objective and may therefore remain fragile under long-horizon or dynamically evolving conditions.

In this work, we analyze the distinction between observation-space sequence modeling and latent-state world modeling. Building on this analysis, we introduce the perspective of Latent Dynamics Inference (LDI), which interprets language as a partial and lossy observation channel of underlying environment dynamics rather than as isolated symbolic sequences. Under this view, learning can be interpreted as a cross-space inference problem in which models recover latent states and transition structure from observational data.

To empirically investigate this hypothesis, we introduce \textsc{Flux}, a novel sequential reasoning environment designed to compare observation-space reasoning with explicit latent-state planning. We evaluate frontier LLMs against reinforcement-learning agents operating directly over a formal world model extracted from natural-language rules, enabling controlled analysis of dynamic reasoning, state tracking, and long-horizon planning behavior.

The primary contributions of this work are as follows:

\begin{itemize}
    \item We formalize the distinction between autoregressive sequence modeling and latent-state world modeling as an objective-level mismatch between observation prediction and environment dynamics modeling.

    \item We analyze the structural implications of this mismatch for hallucination, causal reasoning, long-horizon consistency, and grounded decision-making.
    
    \item We introduce Latent Dynamics Inference (LDI), a conceptual perspective that interprets language and multimodal observations as indirect evidence of latent state trajectories and causal dynamics.
    
    \item We present \textsc{Flux}, a novel sequential game environment specified entirely through natural-language rules, designed to evaluate latent-state reasoning and long-horizon planning.

    \item We present a controlled case study using \textsc{Flux} to illustrate, as a proof of concept, that latent structure extracted from natural-language observations supports more stable state-tracking and planning behavior than observation-space reasoning alone.
\end{itemize}

\section{Related Work}

LLMs have driven a paradigm shift in artificial intelligence, achieving remarkable success in text generation, code synthesis, and knowledge-intensive reasoning tasks~\cite{raza2025industrial}. However, this progress has not translated equally to domains where intelligence is fundamentally grounded in interaction with dynamic environments. Fields such as robotics, industrial automation, autonomous driving, and surgical decision support remain only marginally improved by purely language-based models. Tasks including real-time traffic control, embodied manipulation, adaptive control systems, and AI-assisted surgery require continuous state estimation, causal reasoning, and long-horizon planning—capabilities that are not naturally supported by next-token prediction objectives~\cite{wu2024mldt,malathi2025ai}.

This gap has motivated a growing body of research investigating the limitations of LLMs beyond static text-based benchmarks. Hallucination, for instance, has been widely documented as a systemic issue, where models generate coherent yet factually incorrect outputs. Such behavior is increasingly understood as a consequence of optimizing for statistical likelihood without enforcing grounding or factual consistency. Similarly, recent studies in causal reasoning demonstrate that while LLMs can approximate observational correlations, they struggle with interventional and counterfactual reasoning, limiting their applicability in decision-critical settings~\cite{bhasuran2026evaluation}.

In parallel, the reinforcement learning (RL) and embodied AI communities have long emphasized the importance of world models for intelligent behavior. Early work by Ha and Schmidhuber introduced the concept of learning compact latent representations of environments to enable simulation and control~\cite{ha2018world}. Building on this idea, model-based RL approaches such as Dreamer and its extensions (DreamerV2, DreamerV3) have demonstrated that agents can learn latent dynamics models and perform planning entirely within imagined trajectories~\cite{okada2022dreamingv2,li2025dreamerv3}. These approaches highlight the importance of explicitly modeling state transitions rather than relying solely on observed data distributions.

Further advances include PlaNet, MuZero, and related architectures, which integrate representation learning, dynamics modeling, and planning into unified frameworks~\cite{koul2020dream,guei2025demystifying}. MuZero, in particular, showed that it is possible to learn environment dynamics implicitly while still achieving strong performance in complex domains such as Atari and Go. Collectively, these methods demonstrate that learning and exploiting transition dynamics is central to effective decision-making in complex environments.

More recently, there has been growing interest in combining LLMs with planning and search techniques. Methods such as Tree-of-Thought prompting, ReAct, and LLM-guided Monte Carlo Tree Search attempt to augment language models with structured reasoning capabilities~\cite{long2023large,mo2024tree,yao2022react,zhou2023language}.  While these approaches improve performance in structured reasoning tasks, they largely rely on external scaffolding rather than learning explicit environment dynamics.

Efforts to bridge this gap have also emerged in multimodal and embodied AI systems, where language is integrated with perception and action. However, most existing approaches treat language as an interface rather than a substrate for learning environment structure. Consequently, they underutilize the potential of language as a source of structured information about environment dynamics.

In contrast to prior work, this paper explores the hypothesis that language can serve not merely as an interface or output modality, but as a structured source of supervision for learning environment dynamics. We formalize this view in the following sections by interpreting textual data as indirect observations of latent state transitions.

\section{Formal Analysis: Sequence Modeling vs. World Modeling}
\subsection{Sequence Modeling vs. World Modeling}

Let $D$ denote a dataset of token sequences. Autoregressive language models are trained to maximize the expected log-likelihood:

\begin{equation}
\max_{\theta} \; \mathbb{E}_{x \sim D} \sum_{t} \log P_{\theta}(x_t \mid x_{<t})
\end{equation}

This objective factorizes the joint distribution over sequences into conditional token probabilities, enabling efficient learning of linguistic regularities. While effective for modeling linguistic structure, this objective operates purely over observed sequences and does not explicitly enforce consistency with underlying generative processes.

In contrast, a world model aims to learn the underlying dynamics of an environment:

\begin{equation}
s_{t+1} = f_{\theta}(s_t, a_t), \quad a_t \sim \pi(a_t \mid s_t)
\end{equation}

where $s_t \in \mathcal{S}$ denotes the latent state, $a_t \in \mathcal{A}$ denotes an action, $f_{\theta}$ defines transition dynamics, and $\pi$ is a policy. This formulation models how states evolve under interventions, enabling prediction, simulation, and planning within a structured environment.

The distinction between these objectives reflects a fundamental mismatch in what is being modeled:

\begin{itemize}
    \item Autoregressive models approximate distributions over observations (token sequences),
    \item World models approximate dynamics over latent states (environment configurations).
\end{itemize}

This difference has important implications. In particular, high sequence likelihood does not necessarily imply causal consistency, and accurate token prediction does not inherently correspond to modeling underlying state transitions. We hypothesize that part of this mismatch arises from the learning objective itself rather than solely from limitations in scale or data. As a result, even highly accurate sequence models may fail to support reasoning under interventions, feedback, or long-horizon dependencies, where explicit modeling of state evolution is required.

It is important to note that LLMs are not entirely devoid of latent structure. Probing studies have demonstrated that LLMs internally represent entity states, spatial relations, and elementary physical regularities~\cite{li2022emergent,karvonen2024emergent}. However, the key distinction is that there exists no objective forcing this structure to be causally consistent, compositionally updatable, or aligned with transition dynamics. The mismatch is therefore not binary—it is a matter of structural alignment: LLMs may incidentally approximate world-model behavior in narrow, well-represented domains, but such behavior remains fragile, non-systematic, and unguaranteed.

\subsection{Observation Space vs Latent State Space}

This objective mismatch can be further analyzed through the distinction between representation space and latent state space.

LLMs operate over a token-based representation space $\mathcal{X}$, where sequences are mapped into high-dimensional embeddings via $\phi: \mathcal{X} \rightarrow \mathbb{R}^d$. The model learns statistical dependencies within this space:

\begin{equation}
P(x_t \mid x_{<t}) = P\big(\phi(x_t) \mid \phi(x_{<t})\big)
\end{equation}

Thus, reasoning in LLMs is performed over relations between token representations, capturing co-occurrence patterns and latent semantics, but not necessarily the underlying generative processes.

In contrast, world models operate over a latent state space $\mathcal{S}$ that encodes entities, properties, and interactions. The transition function models how states evolve under actions, enabling reasoning over causal dynamics as follow:

\begin{equation}
s_{t+1} = f_{\theta}(s_t, a_t)
\end{equation}

The token space $\mathcal{X}$ can be interpreted as a projection of the latent state space $\mathcal{S}$. Formally, there exists a (generally unknown and lossy) observation mapping:

\begin{equation}
x_t \sim P(x_t \mid s_t)
\end{equation}

which generates textual descriptions from underlying states.

\begin{figure}[t]
    \centering
    \includegraphics[width=\linewidth]{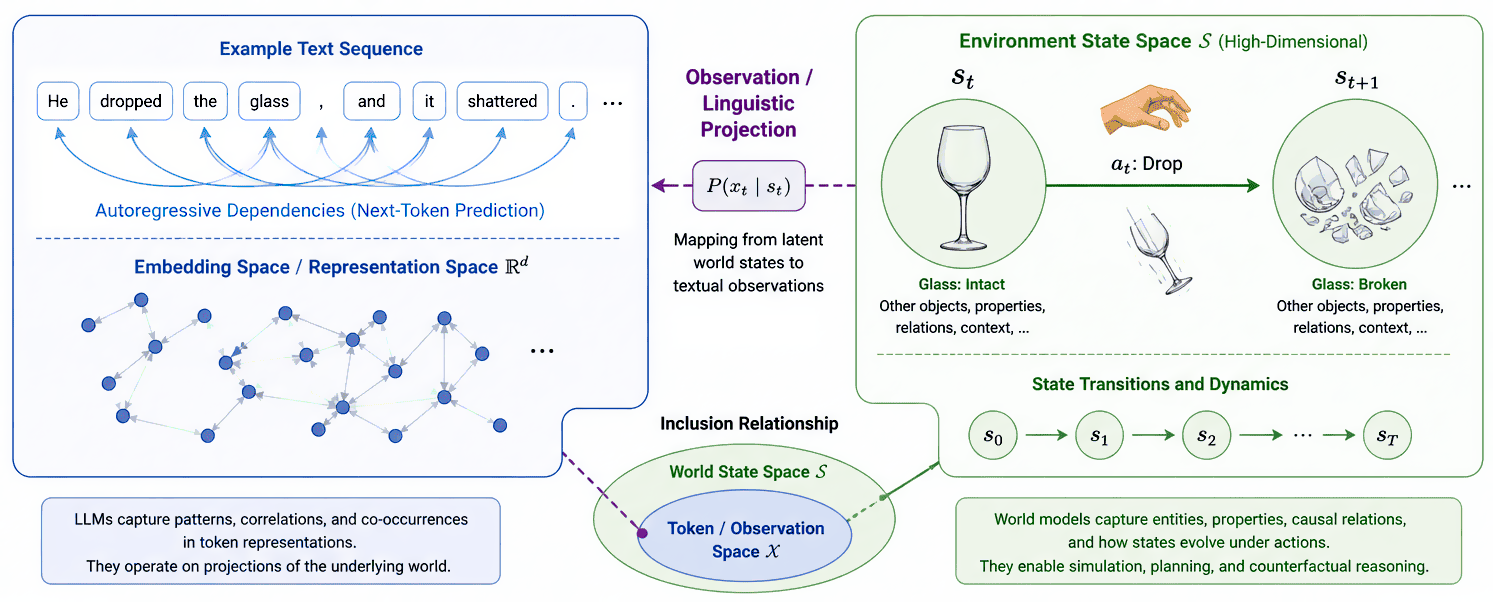}
    \caption{
    Token representation space $\mathcal{X}$ versus latent world state space $\mathcal{S}$. 
    LLMs model statistical dependencies in $\mathcal{X}$, while world models capture causal dynamics in $\mathcal{S}$. 
    The mapping $P(x_t \mid s_t)$ represents a lossy projection from states to observations.
    }
    \label{fig:token_vs_state}
\end{figure}

As illustrated in Figure~\ref{fig:token_vs_state}, LLMs operate in the observation space $\mathcal{X}$, while world models operate in the latent state space $\mathcal{S}$, capturing dynamics driven by actions.

This induces a hierarchy:

\begin{itemize}
    \item \textbf{Observation Space ($\mathcal{X}$):} Encodes linguistic representations and statistical relationships,
    \item \textbf{State Space ($\mathcal{S}$):} Encodes causal structure and environment dynamics.
\end{itemize}

Because $\mathcal{X}$ is a compressed and partial view of $\mathcal{S}$, modeling relationships in $\mathcal{X}$ alone may be insufficient to reliably recover environment dynamics in settings requiring persistent state tracking and intervention reasoning. This perspective motivates learning directions that explicitly infer and evolve latent states, rather than relying solely on autoregressive sequence modeling.

A further challenge concerns the identifiability of the inverse mapping $P(s \mid x)$. Because the observation process $P(x \mid s)$ is many-to-many—multiple distinct latent states can generate similar linguistic expressions—the recovery of latent dynamics from text alone is fundamentally under-determined. This is not merely a practical difficulty but a structural one: additional data alone may not fully resolve this ambiguity without auxiliary constraints or inductive biases that privilege certain state representations over others. Addressing this identifiability problem appears to be a central challenge for any framework that aims to learn world models from language.

\subsection{ Structural Limitations of Next-Token Prediction}

These limitations may reflect deeper properties of autoregressive training objectives beyond simple scaling constraints. By framing the shortcomings of LLMs against the architectural strengths of world models, we can identify a pathway toward more robust artificial intelligence.

\begin{table}[t]
\centering
\caption{
Conceptual comparison between autoregressive sequence models and latent world models across core properties related to reasoning, planning, and state representation.
}
\label{tab:sequence_vs_world_models}

\renewcommand{\arraystretch}{1.25}

\begin{tabular}{
>{\raggedright\arraybackslash}p{0.33\linewidth}
>{\raggedright\arraybackslash}p{0.28\linewidth}
>{\raggedright\arraybackslash}p{0.28\linewidth}
}
\hline
\rowcolor{headergray}
\textbf{Capability} & \textbf{Sequence Models} & \textbf{World Models} \\
\hline

\rowcolor{white}
Observation prediction & Strong & Secondary objective \\

\rowcolor{headergray}
Intervention reasoning & Implicit / weak & Explicitly modeled \\

\rowcolor{white}
Long-horizon consistency & Error accumulation & State-based stabilization \\

\rowcolor{headergray}
Counterfactual simulation & Approximate / emergent & Structurally supported \\

\rowcolor{white}
State persistence & Implicit / transient & Persistent latent states \\

\rowcolor{headergray}
Grounded environment dynamics & Indirect & Explicit transition modeling \\

\rowcolor{white}
Feedback-based correction & Limited & Closed-loop correction \\
\hline

\end{tabular}
\end{table}

Table~\ref{tab:sequence_vs_world_models} summarizes the principal structural differences between autoregressive sequence models and latent world models across key dimensions related to reasoning, planning, and environment interaction.

\subsubsection{Objective-Induced Hallucination vs. Latent Grounding}
Hallucination in LLMs is often framed as a "bug," but it is more accurately a feature of the autoregressive objective. Since the model is optimized to minimize cross-entropy over a token distribution $P(x_t \mid x_{<t})$, it prioritizes \textit{statistical plausibility} over \textit{factual grounding}. When a model enters an out-of-distribution state, it drifts toward high-probability "average" sequences that may be semantically hollow or factually incorrect.

World models address this differently by distinguishing two failure modes that LLM hallucination conflates: internal inconsistency and factual incorrectness. A world model grounds generation in an explicit latent state $s_t$, which means its outputs are constrained to be consistent with the model's internal representation—regardless of whether that representation is accurate. Crucially, this internal consistency makes errors detectable: when predicted transitions diverge from observed evidence, the closed-loop feedback mechanism can trigger correction. In contrast, hallucinations in LLMs are more difficult to detect internally when no explicit state representation is maintained against which generated outputs can be checked. The primary advantage of world-model architectures is therefore not immunity to error, but the capacity for closed-loop correction through explicit state tracking and transition consistency.

\subsubsection{Observational Correlation vs. Causal Intervention}
A fundamental barrier for LLMs is the distinction between \textit{seeing} and \textit{doing}. LLMs are trained on passive, static corpora, capturing observational correlations $P(Y \mid X)$. However, true intelligence requires understanding interventional dynamics $P(Y \mid \text{do}(X))$, where an agent predicts the outcome of its own actions~\cite{bhasuran2026evaluation, ge2026review}. Without this causal lens, LLMs struggle with counterfactual reasoning—the ability to imagine how the environment would change if a single variable were altered.

World models are structurally more aligned with intervention-based reasoning. By learning a transition function $f_{\theta}(s_t, a_t)$, they encode the action $a_t$ as an explicit input to the transition, which is conceptually aligned with intervention-based reasoning frameworks such as Pearl's do-operator: the action is an intervention on the current state, not a correlated observation. This formal alignment between $f_{\theta}(s_t, a_t)$ and the interventional distribution $P(s_{t+1} \mid \mathrm{do}(a_t), s_t)$ is the key structural property that distinguishes world models from sequence models, and it is present by construction rather than by emergence.

However, this structural property does not guarantee causal correctness. A world model trained only on passive observational data may still learn spurious correlations rather than true causal mechanisms. Causal reasoning therefore depends not only on architecture, but also on sufficient interventional signals during training.

While an LLM can only predict the next word in a description of a fire, a world model can simulate the heat, the oxygen depletion, and the resulting structural collapse, allowing for reasoning that is grounded in the laws of the environment rather than the patterns of the text.

\subsubsection{Autoregressive Divergence vs. Stable State Tracking}
The instability of LLMs in long-horizon tasks—such as complex planning or multi-step math—arises from compounding error. In a pure autoregressive setup, each token generation is conditioned on its own previous (potentially erroneous) outputs. As the sequence length $n$ increases, the probability of remaining within a valid reasoning trajectory decays exponentially ($p^n$), leading to "logical incoherence" in extended interactions.

World-model architectures are structurally better positioned to address this through closed-loop feedback and state tracking. Instead of relying on a burgeoning history of tokens (which may contain errors), a world model updates a fixed-dimensional latent state $s_t$. This state acts as an anchor, allowing the system to assimilate feedback from the environment and correct its internal representation. This transition from open-loop sequence generation to closed-loop state estimation appears important for tasks requiring long-term consistency, such as autonomous driving or surgical robotics, where the "cost" of a single divergent error is catastrophic.

\subsubsection{Synthesis: Language as a Trace of World Dynamics}
The path forward lies in reinterpreting natural language not as the \textit{goal} of learning, but as a \textit{sensor} for the world. If we view textual corpora as a lossy projection of underlying environment states (as argued in Section 3.2), then LLMs can be seen as highly efficient encoders of world knowledge that currently lack the "physics engine" to execute that knowledge. Integrating the vast, diverse knowledge encoded in LLMs with the causal, stateful structure of world models offers a dual-process system: one that possesses the breadth of human language and the depth of physical and logical reality.

\section{Latent Dynamics Inference (LDI)}\label{sec:ldi}

\subsection{Can World Models Be Learned from Language?}

We posit the following hypothesis:

\begin{quote}
\textit{Natural language corpora constitute implicit, partial observations of 
trajectories arising from interactions between agents and environments, encoded 
through linguistic representations.}
\end{quote}

This perspective reframes text not as static symbolic data, but as a compressed 
and indirect record of dynamic processes. Rather than treating language as an end 
product, we interpret it as an observation channel through which underlying state 
transitions and actions are indirectly expressed.

Under this formulation, textual sequences can be interpreted as observations 
generated from an underlying latent dynamical system. Specifically, a sequence of 
tokens $x_{1:T}$ is assumed to arise from a corresponding (unobserved) trajectory:

\begin{equation}
(s_1, a_1, s_2, a_2, \dots, s_T), \quad s_t \in \mathcal{S}, \; a_t \in \mathcal{A}
\end{equation}

where $s_t$ denotes the latent state of the environment and $a_t$ denotes an action 
or event inducing a transition. The observed tokens are generated via a stochastic 
observation process:

\begin{equation}
x_t \sim P(x_t \mid s_t)
\end{equation}

As discussed in Section 3, this observation process provides only partial access to 
the underlying state, since multiple latent configurations may correspond to similar 
linguistic expressions.

Within this perspective:
\begin{itemize}
    \item Tokens correspond to noisy and partial observations of latent states.
    \item Sentences encode localized transitions or events.
    \item Discourse and narratives capture extended trajectories over time.
\end{itemize}

Consider the sentence:

\begin{quote}
``He dropped the glass, and it shattered.''
\end{quote}

This linguistic observation can be mapped to a latent transition:

\begin{itemize}
    \item $s_t$: \texttt{Glass(Intact), Held=True}
    \item $a_t$: \texttt{Release / Drop}
    \item $s_{t+1}$: \texttt{Glass(Broken), OnFloor=True}
\end{itemize}

Importantly, this perspective implies a shift in what is being modeled. Traditional 
language modeling primarily learns relationships within the representation space 
$\mathcal{X}$ itself, capturing statistical dependencies, semantic associations, and 
structural patterns between tokens and sequences. In this setting, reasoning is 
performed largely within a self-contained linguistic manifold, where the objective 
is to predict observations from other observations.

In contrast, learning world models from text requires moving beyond intra-linguistic 
relationships toward modeling the correspondence between two distinct spaces: the 
observation space $\mathcal{X}$ and the latent environment space $\mathcal{S}$. The 
central challenge is therefore not only to model relationships between tokens, but 
to infer which aspects of textual structure reflect meaningful latent states, 
transitions, causal mechanisms, and interactions within the underlying environment.

Under this view, language becomes a compressed interface to world dynamics. Textual 
corpora encode partial evidence about physical processes, social interactions, 
intentions, beliefs, and temporal evolution, even though these variables are not 
directly observed. Consequently, exploiting text for world modeling requires 
identifying and recovering the latent structure that gives rise to linguistic 
observations.

This reframes learning as a cross-space inference problem:

\begin{equation}
\mathcal{X} \rightarrow \mathcal{S}
\end{equation}

We refer to this perspective as \textbf{Latent Dynamics Inference (LDI)}: the process 
of inferring hidden environment states, transition structures, and causal dynamics 
from incomplete observational sequences such as language, video, or multimodal 
interaction traces. From the perspective of LDI, intelligence may depend less on predicting observations themselves and more on inferring the latent processes that generate them, but the recovery and evolution of the latent dynamical 
processes that generate those observations. Language modeling therefore becomes a 
special case of a broader inference problem in which observable sequences provide 
indirect evidence about hidden state trajectories.

A key question within LDI concerns the structure of what language systematically 
encodes and what it systematically omits. We identify three classes of information 
that are reliably underrepresented in textual observations: (i) metric and 
quantitative properties (e.g., exact velocities, distances, or forces), which are 
rarely expressed with precision in natural language; (ii) simultaneous and concurrent 
events, which are difficult to express in the linear sequence structure of text; and 
(iii) private or first-person perceptual states, which are accessible only through 
indirect report. Conversely, language reliably encodes causal relationships, 
sequential structure, intentionality, and normative context. This asymmetry implies 
that text-derived world models will be stronger at modeling abstract, relational, and 
causal dynamics, and weaker at modeling precise quantitative or perceptual 
dynamics—a constraint that should inform architectural choices under LDI.

\begin{figure}[t]
    \centering
    \includegraphics[width=0.8\linewidth]{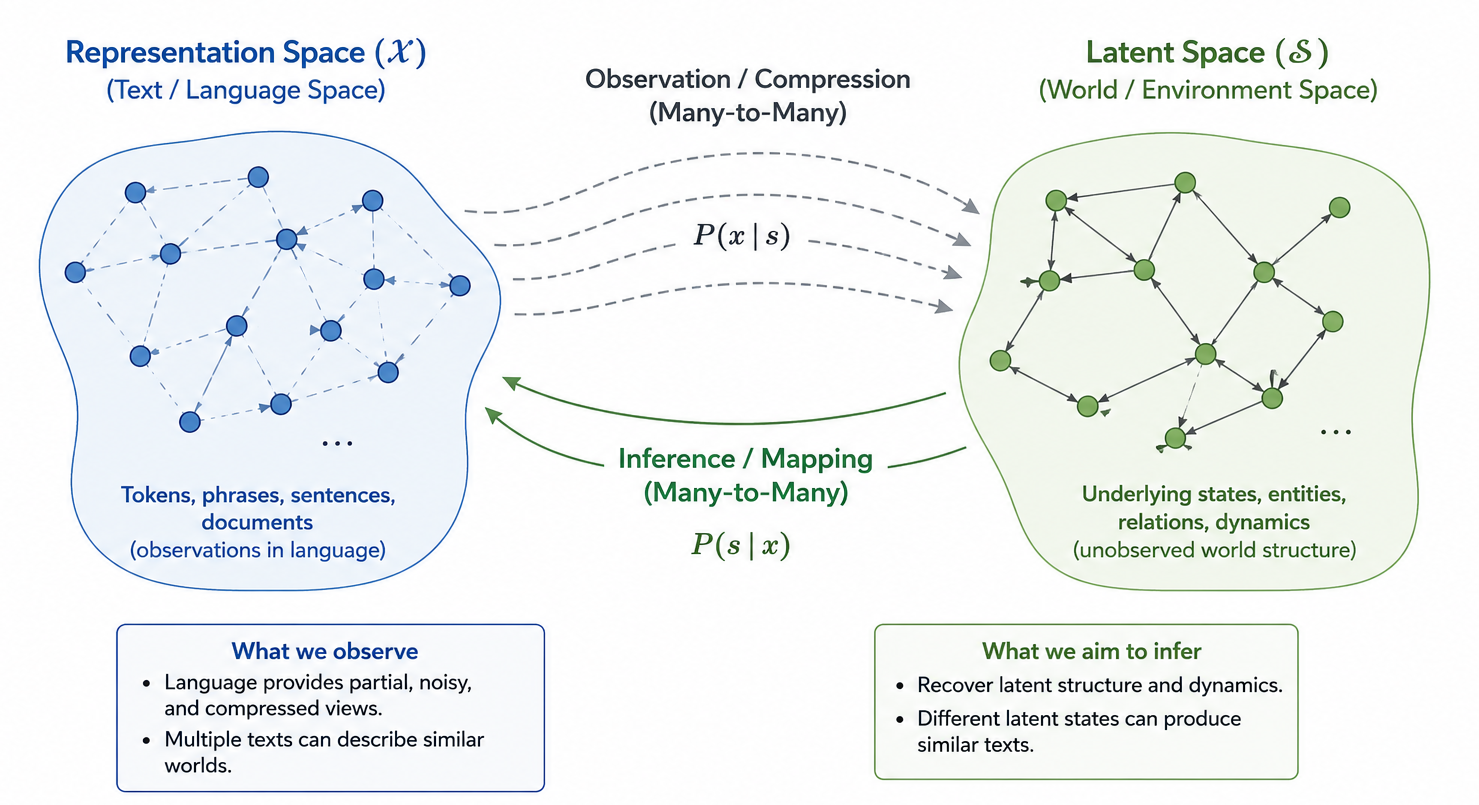}
    \caption{
Relationship between the linguistic representation space $\mathcal{X}$ and the 
latent world state space $\mathcal{S}$ under the Latent Dynamics Inference (LDI) 
perspective. Language observations are generated through a lossy many-to-many mapping 
$P(x \mid s)$, while LDI aims to infer latent states and dynamics through the 
inverse mapping $P(s \mid x)$. Unlike traditional language models that learn 
relationships within $\mathcal{X}$, LDI seeks to recover causal and dynamical 
structure from compressed textual observations.
}
    \label{fig:x_to_s_mapping}
\end{figure}

Figure~\ref{fig:x_to_s_mapping} illustrates the proposed LDI perspective between the 
linguistic representation space $\mathcal{X}$ and the latent world state space 
$\mathcal{S}$.

Crucially, a significant portion of human knowledge is represented textually, 
including scientific discoveries, procedural reasoning, social behavior, historical 
events, and descriptions of physical interactions. Although language provides only 
partial and lossy access to the underlying world state, its scale and diversity make 
it a uniquely rich source of supervision for learning latent environment structure 
under LDI. The challenge is therefore developing mechanisms capable of extracting, 
organizing, and evolving the latent dynamical information implicitly encoded within 
textual data.

This perspective suggests three design principles for systems that must reason over dynamic environments. First, observations should be decoded into explicit latent states before reasoning is performed, rather than treating raw observations as the substrate for inference. Second, state representations should be updated through transition functions rather than accumulated conversational context. Third, planning and decision-making should occur within state space rather than observation space. These principles should be interpreted as diagnostic hypotheses rather than established requirements for intelligence: systems that violate them---as autoregressive LLMs do by construction---are predicted to exhibit degraded performance precisely in tasks requiring persistent state tracking and multi-step planning. The experiments in Section~5 are designed to test this prediction directly, by comparing agents that operate in $\mathcal{X}$ against agents that operate in $\mathcal{S}$.

\subsection{Multimodal Extensions: Learning from Visual and Temporal Observations}

The above perspective naturally extends beyond textual corpora to other forms of sequential data, particularly video and animation, which provide rich observations of environment dynamics.

While visual data is represented at the level of pixels, these should be interpreted as observations rather than primary units of reasoning.

Formally, a video sequence $v_{1:T}$ can be viewed as:

\begin{equation}
v_t \sim P(v_t \mid s_t)
\end{equation}

where $s_t$ represents the latent physical state of the environment (e.g., object positions, velocities, forces), and $v_t$ is the corresponding rendered frame. As in the textual case, observations provide only indirect access to the underlying dynamics.

For example, consider a simple physical process such as a bouncing ball. At the pixel level, the video consists of changing color intensities across frames. However, a more meaningful interpretation is:

\begin{itemize}
    \item $s_t$: \texttt{Ball(Position, Velocity), Gravity, SurfaceProperties}
    \item $a_t$: \texttt{Collision / External Forces}
    \item $s_{t+1}$: Updated physical state following dynamics (e.g., velocity inversion upon impact)
\end{itemize}

The apparent motion in pixel space is governed by underlying physical laws rather than raw visual transitions. Learning directly in pixel space obscures these dynamics, whereas learning in state space enables generalization, prediction, and simulation.

This observation reinforces a broader principle:

\begin{itemize}
    \item \textbf{Text} provides access to implicit \textit{abstract} trajectories (e.g., social, cognitive, and causal processes),
    \item \textbf{Video} provides access to implicit \textit{multi-level trajectories}, including physical dynamics (motion, interaction), as well as \textit{behavioral, social, and affective states} (e.g., gestures, facial expressions, emotional reactions, and interpersonal interactions).
\end{itemize}

Both modalities can be treated as complementary observation channels of a shared latent dynamical system.

Consequently, large-scale multimodal data (text, video, simulation traces) can be viewed as a complementary and unified source of supervision for world modeling. While text encodes high-level abstractions such as intent, belief, and causality, video encodes low-level physical regularities and temporal continuity, as well as higher-level behavioral, social, and affective cues expressed through motion, interaction patterns, and visual context.

This suggests one possible direction for progress: learning shared latent representations that align these observation channels, rather than scaling modalities independently. Such representations would enable models to reason simultaneously about physical processes, social interactions, and cognitive states—bridging perception, language, and action within a unified dynamical framework.

\section{Experimental Design and Results}
\label{sec:experiments}

The experimental framework is not intended to establish a general superiority claim between large language models and world-model-based agents. Instead, it serves as a controlled case study illustrating a central implication of the LDI perspective introduced in Section~3: that latent state structure can be inferred from natural-language observations and compiled into an explicit state-transition system. Specifically, we investigate whether transition dynamics, action effects, and terminal conditions can be extracted from a purely textual game specification and mapped from the observation space $\mathcal{X}$ into a latent state space $\mathcal{S}$.

To examine this, we construct an environment in which the game rules are provided entirely in natural language, representing the observation space $\mathcal{X}$. These textual descriptions are then systematically transformed into a formal simulator operating in latent space $\mathcal{S}$. This setup enables a separation between two capabilities: (1) structural inference, where rules and dynamics are extracted from language and compiled into an explicit state representation, and (2) dynamic reasoning, where agents must plan and act over time using either observation-only access or direct access to the induced transition dynamics.

The purpose of the evaluation is therefore illustrative rather than benchmark-oriented. The results should not be interpreted as evidence that world-model agents universally outperform autoregressive language models, nor as proof of an inherent limitation of sequence modeling. Rather, the experiment demonstrates a concrete instance of the $\mathcal{X} \rightarrow \mathcal{S}$ mapping central to LDI: that latent transition structure can be induced from textual observations, and that operating directly within this extracted state space yields behavior more consistent with stable state tracking and multi-step planning.

Within this experiment, we compare two classes of agents. LLM-based agents operate purely in $\mathcal{X}$, receiving natural-language state descriptions and selecting actions via next-token prediction. In contrast, reinforcement learning agents operate directly in $\mathcal{S}$, interacting with a deterministic simulator defined by the extracted transition rules and learning policies through experience in the underlying dynamics.

This design enables a controlled comparison between reasoning over observation sequences and reasoning over explicit latent state evolution, even though both agent classes originate from the same natural-language specification.

\begin{figure}[t]
    \centering
    \includegraphics[width=0.95\linewidth]{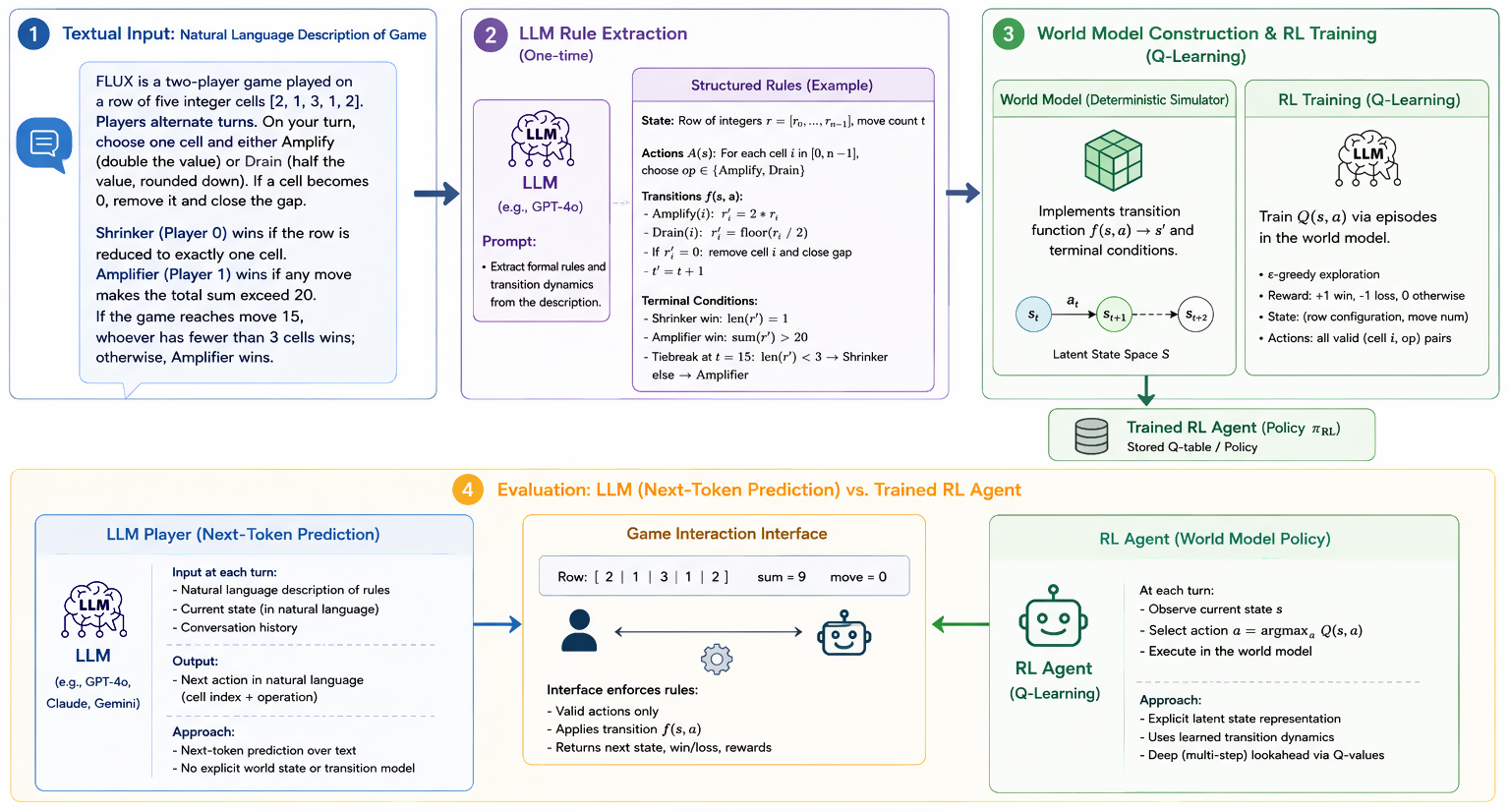}
\caption{Overall experimental framework. A natural language game description is transformed into structured rules using an LLM, which define the environment for training a Q-learning agent. LLMs are later evaluated in a turn-by-turn gameplay setting for comparison with the RL agent.}
    \label{fig:experiment_pipeline}
\end{figure}

The experimental pipeline proceeds in four stages. First, a natural-language description of the game is provided, specifying rules, constraints, and objectives. Second, an LLM is used to compile these rules into a structured executable representation defining the environment dynamics. Third, a Q-learning agent is trained within this induced environment to learn an optimal policy through interaction with the transition system. Finally, LLM-based agents are evaluated in the same environment in a turn-by-turn setting, enabling direct comparison between implicit sequence-based reasoning and explicit world-model-based planning. The full experimental architecture is illustrated in Figure~\ref{fig:experiment_pipeline}.

In this setup, the LLM used for rule extraction plays a clearly scoped role as a one-time compiler that converts natural-language descriptions into a formal specification of the environment. This step corresponds to offline information extraction, where the underlying rules are already explicitly encoded in the input text and the model functions primarily as a translator from natural language into a structured representation of the system dynamics. By isolating rule parsing from execution, this design ensures that any performance differences observed during gameplay cannot be attributed to ambiguity or inconsistency in the environment definition.

Importantly, this usage does not contradict the central hypothesis regarding the limitations of LLMs in constructing latent world models from observation space. The successful compilation of \textsc{Flux} from textual rules demonstrates that the mapping from observation space $X$ to latent state space $S$ is feasible through language modeling. However, the LLM is not engaged here as a dynamic reasoning or planning agent, but rather as a static compiler operating once at initialization. The primary challenge therefore arises during sequential execution under evolving state dynamics, where agents must maintain persistent state tracking and perform multi-step causal reasoning over time. The key distinction is thus not whether LLMs can infer latent structure from language, but whether they can sustain a coherent and persistently updated state-transition process during long-horizon interaction in the absence of an explicit world model.

\subsection{Game Environment: FLUX}

To empirically validate the central thesis---that the fundamental limitation of autoregressive models lies in operating over observation space $\mathcal{X}$ rather than latent environment dynamics $\mathcal{S}$, and that explicit state-based modeling yields superior performance in sequential decision-making tasks---we designed a novel two-player game, \textsc{Flux}, specifically constructed to satisfy four criteria: (i) rules expressible in a natural-language paragraph, ensuring no prior exposure in any LLM training corpus; (ii) fully observable, sequential state transitions amenable to conversational LLM play; (iii) genuine planning depth that defeats pure pattern-matching; and (iv) asymmetric win conditions creating strategic tension across all time horizons.

\begin{tcolorbox}[colback=lightyellow, colframe=black, boxrule=0.4pt, arc=2mm, left=2mm, right=2mm, top=2mm, bottom=2mm]
\textsc{Flux} is played on a row of five integer cells, initialised as $[2, 1, 3, 1, 2]$. 

Players alternate turns; on each turn a player selects one cell and applies either \textsc{Amplify} ($\times 2$) or \textsc{Drain} ($\lfloor \div 2 \rfloor$). Any cell reaching zero is removed and the row contracts. 

Two terminal conditions exist: (a)~the \emph{Shrinker} (Player~0) wins immediately if the row is reduced to exactly one cell; (b)~the \emph{Amplifier} (Player~1) wins immediately if any move causes the total sum to exceed 20.

A tiebreak at move~15 awards victory to the Shrinker if fewer than three cells remain, otherwise to the Amplifier.
\end{tcolorbox}

The formal world model extracted from this natural-language description constitutes the latent state space $\mathcal{S} = (\mathbf{c}, t)$, where $\mathbf{c} \in \mathbb{Z}_{>0}^{n}$ is the current cell vector and $t \in \mathbb{N}$ is the move counter. The transition function $f_\theta : \mathcal{S} \times \mathcal{A} \to \mathcal{S} \times \Omega$ is deterministic and causally grounded; it is implemented as an explicit Python simulator with zero statistical approximation.

\subsection{Agents}

Four agent types were instantiated, representing increasing levels of latent state access:

\textbf{Random Agent.} Selects uniformly at random from the set of valid actions. Serves as the lower-bound baseline.

\textbf{Heuristic Agent. } Implements one-step greedy look-ahead using hand-coded heuristics derived solely from the natural-language rule text---no access to the formal world model. The Heuristic Shrinker scores candidate moves by the reduction in row length minus a sum penalty; the Heuristic Amplifier maximises sum growth and row preservation. 

\textbf{RL Agent (World Model).} A tabular Q-learning agent trained entirely within the extracted world model simulator. The state key is $({\rm cells}, {\rm move\_num})$; actions are encoded as $2i + \mathbb{1}[{\rm op}={\rm drain}]$. Hyperparameters: $\alpha = 0.2$, $\gamma = 0.92$, $\varepsilon$ decaying from $1.0$ to $0.05$ at rate $0.9997$ per episode. Training employed a cyclic three-mode curriculum over 30{,}000 episodes: (1)~Shrinker vs.\ Random, (2)~Amplifier vs.\ Random, (3)~symmetric self-play. This agent has full access to $\mathcal{S}$ and learns through repeated causal interaction with $f_\theta$.

\textbf{Large Language Model Agent.} Three large-scale instruction-tuned language models were evaluated in prompting mode. At each turn the current state was projected to natural language via $P(\mathbf{x}|\mathbf{s})$ and appended to a running conversation context. The model was prompted to reply with selected operation.

\subsection{Training Convergence}

\begin{figure}[t]
    \centering
    \includegraphics[width=0.95\linewidth]{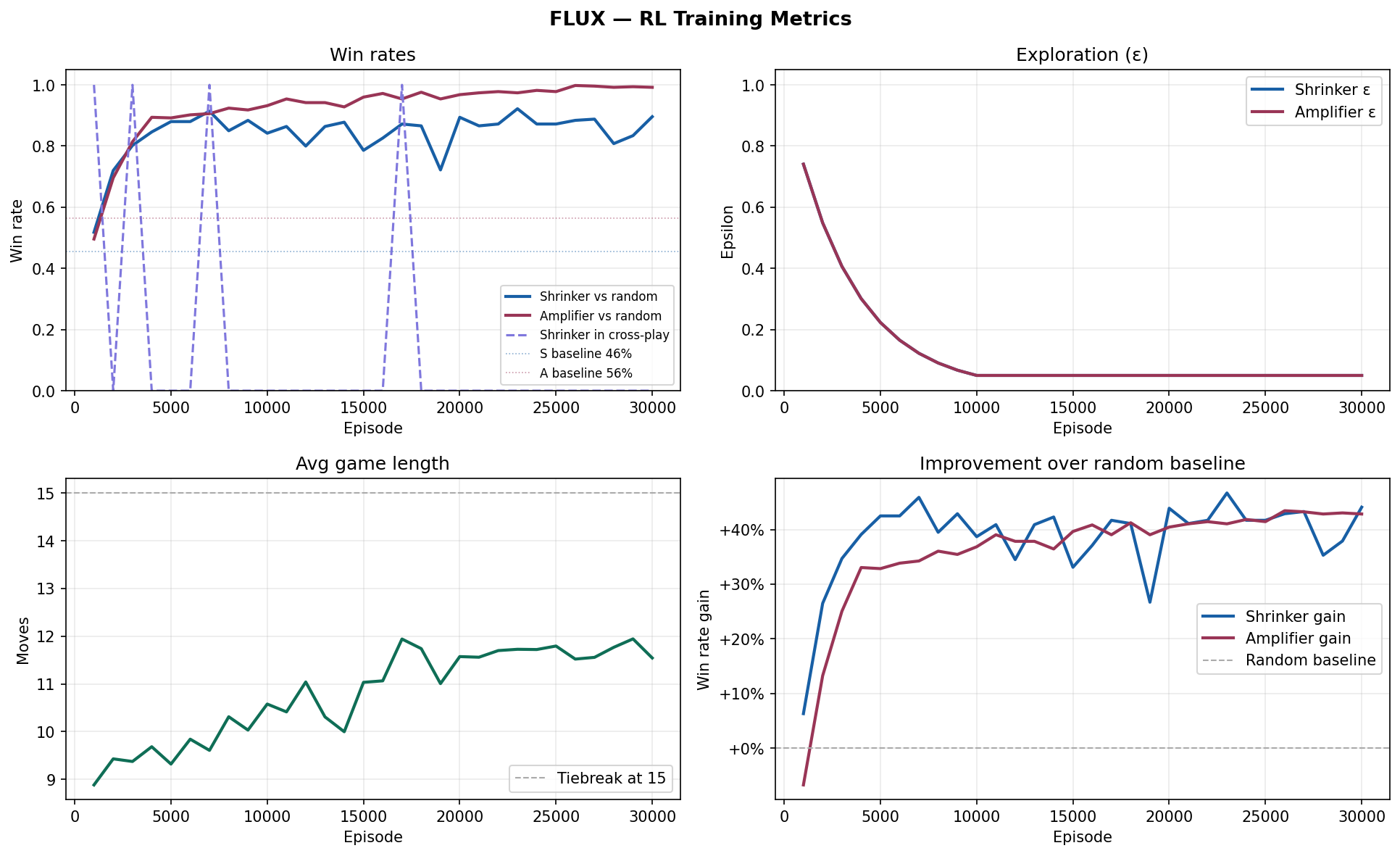}
    \caption{Training dynamics of the Q-learning agents.}
    \label{fig:flux_training}
\end{figure}

After 30{,}000 training episodes, the Shrinker Q-table covered 3{,}092 unique states, while the Amplifier Q-table covered 2{,}657 states, together spanning the reachable region of $\mathcal{S}$ under competitive play. Both $\varepsilon$ values converged to their respective minima, indicating sufficient exploration. The corresponding training dynamics are illustrated in Figure~\ref{fig:flux_training}.

\subsection{World-Model Agent vs.\ Rule-Based Baselines}

\begin{table}[t]
\caption{RL Agent Performance vs.\ Baselines (1{,}000 games each).}
\label{tab:rl_results}
\centering
\begin{tabular}{llrr}
\toprule
\textbf{Matchup} & \textbf{Evaluated Role} & \textbf{Win\%} & \textbf{Avg.\ Moves} \\
\midrule
Random vs.\ Random       & Shrinker  & 43.3\% & 12.3 \\
Heuristic vs.\ Random    & Shrinker  & 77.6\% & 10.2 \\
RL vs.\ Random           & Shrinker  & 89.5\% & 11.1 \\
RL vs.\ Heuristic        & Shrinker  &  0.0\% & 13.0 \\
\midrule
Random vs.\ Random       & Amplifier & 57.4\% & 12.0 \\
Random vs.\ Heuristic    & Amplifier & 99.5\% &  8.8 \\
Random vs.\ RL           & Amplifier & 98.8\% & 11.6 \\
Heuristic vs.\ RL        & Amplifier &100.0\% & 15.0 \\
\bottomrule
\end{tabular}
\end{table}

Table~\ref{tab:rl_results} reports win rates and average game lengths over 1{,}000 episodes per matchup. First, the RL-controlled roles substantially outperform their random-policy counterparts: the Shrinker improves from $43.3\%$ to $89.5\%$ win rate (+46.2 pp), while the Amplifier improves from $57.4\%$ to $98.8\%$ (+41.4 pp), confirming that latent-state learning extracts non-trivial strategic knowledge. Second, the Heuristic agent---despite one-step look-ahead---is comprehensively defeated by the RL Amplifier ($100\%$ loss rate, games consistently forced to tiebreak at move~15). This is consistent with the theoretical prediction that one-step $\mathcal{X}$-space reasoning cannot substitute for multi-step causal planning in $\mathcal{S}$. Third, the symmetric Nash equilibrium visible in the RL vs.\ RL cross-play row ($0\%$ Shrinker wins, all games reaching move~15) indicates that both agents converged to complementary dominant strategies: the Amplifier learned to force the sum constraint to bind exactly at the tiebreak boundary, while the Shrinker could not overcome this without violating the sum cap. This constitutes an emergent equilibrium discovered entirely through self-play within the world model, not through any linguistic prior.

\subsection{Large Language Model vs.\ World Model}

To directly test the $\mathcal{X}$-space vs.\ $\mathcal{S}$-space hypothesis, three LLMs were evaluated against the trained RL agents over 100 games per role (200 games per model). The game description and rules were provided verbatim as a natural-language prompt at the start of each session, together with the board state presented as a formatted table at every turn. Invalid moves---defined as responses failing format validation or specifying an out-of-range index---were recorded separately and counted as forfeited turns (a random legal move was substituted).

\begin{table}[t]
\caption{LLM vs.\ World-Model Agent: \textsc{Flux} Evaluation (100 games per role per model).}
\label{tab:llm_results}
\centering
\renewcommand{\arraystretch}{1.2}
\begin{tabular}{llrrrrr}
\toprule
\multirow{2}{*}{\textbf{LLM}} & \multirow{2}{*}{\textbf{Role}} 
  & \multicolumn{2}{c}{\textbf{Outcome (\%)}} 
  & \multirow{2}{*}{\textbf{Invalid (\%)}} 
  & \multirow{2}{*}{\textbf{Avg.\ Moves}} \\
\cmidrule(lr){3-4}
 & & \textbf{LLM Win} & \textbf{WM Win} & & \\
\midrule
GPT-4o          & Shrinker  &  8\% & 83\% & 9\% & 13.4 \\
GPT-4o          & Amplifier & 14\% & 76\% & 10\% & 11.8 \\
\midrule
DeepSeek-R1  & Shrinker  &  6\% & 80\% & 14\% & 12.9 \\
DeepSeek-R1  & Amplifier & 10\% & 78\% & 12\% & 12.1 \\
\midrule
Gemini 1.5 Pro      & Shrinker  & 14\% & 78\% &  8\% & 13.1 \\
Gemini 1.5 Pro      & Amplifier & 16\% & 79\% &  5\% & 11.6 \\
\midrule
\textbf{Pooled} & \textbf{Both} & \textbf{11.3\%} & \textbf{79.0\%} & \textbf{9.7\%} & \textbf{12.5} \\
\bottomrule
\end{tabular}
\end{table}

As reported in Table~\ref{tab:llm_results}, the world-model RL agent won approximately 79\% of games in aggregate, while LLMs won approximately 11\%, with the remaining $\approx 10\%$ attributable to invalid move forfeits. These results, consistent across both roles and all three models, serve as a case-study illustration of the behavioral difference between agents operating in $\mathcal{S}$ versus $\mathcal{X}$, rather than a general claim about the relative capabilities of world models and LLMs.

Three qualitative failure modes were observed in LLM play. \emph{Sum blindness}: LLMs repeatedly applied Amplify to large cells despite the running sum approaching the cap of 20, triggering the immediate-loss condition; this suggests the model fails to maintain a persistent running sum across turns. \emph{Row-length miscounting}: after several contractions the LLM referenced cell indices that no longer existed, indicating a breakdown in latent row-state tracking. \emph{Horizon myopia}: LLMs occasionally made locally reasonable moves (reducing a cell to zero) without anticipating that the resulting row length would satisfy the Shrinker win condition two turns later.

\begin{figure}[t]
    \centering
    \includegraphics[width=0.95\linewidth]{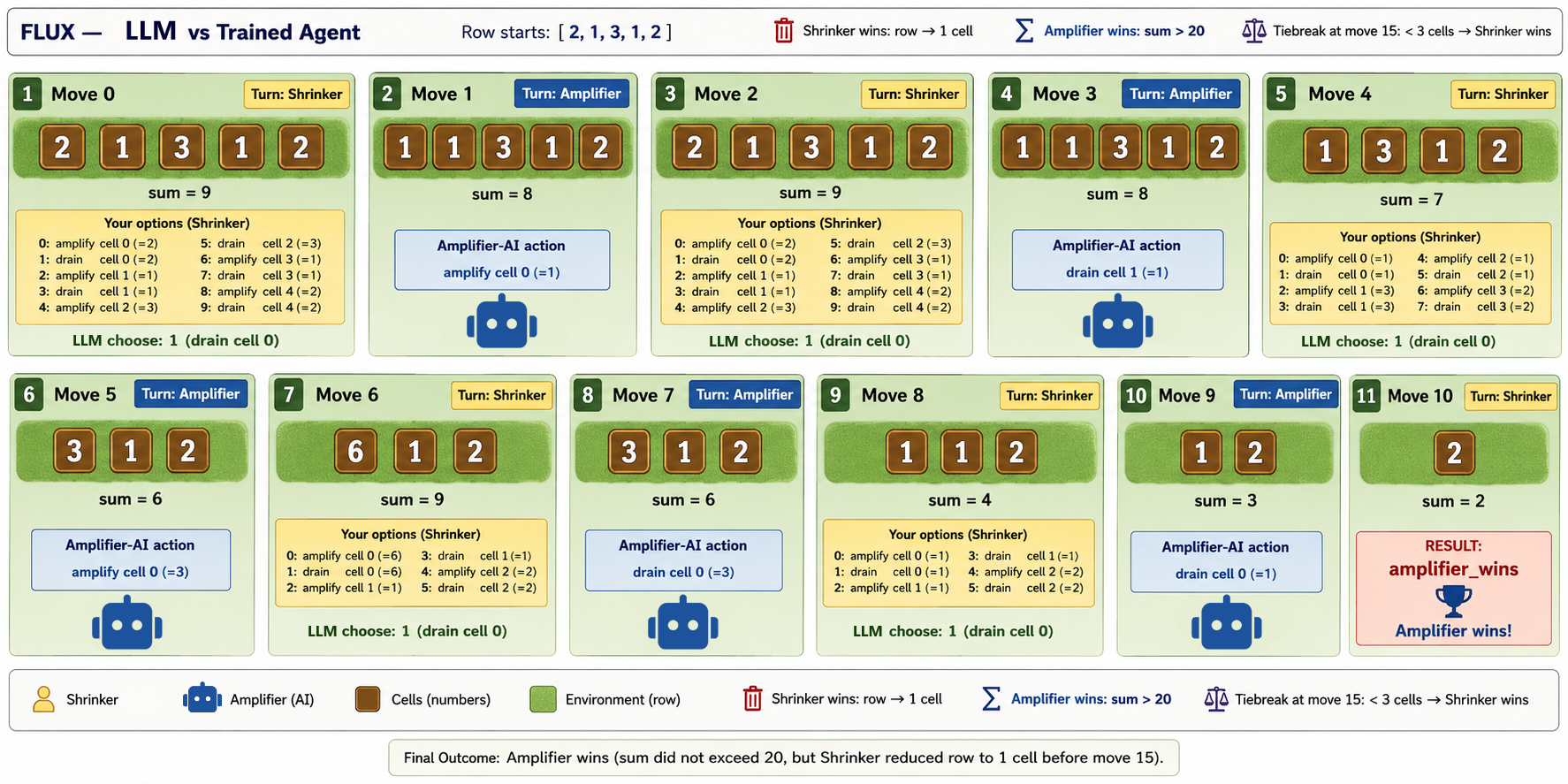}
    \caption{Step-by-step illustration of a full \textsc{Flux} game episode. The figure shows the initial state cells, alternating actions between the LLM-based policy and the world-model RL agent, intermediate state transitions under \textsc{Amplify} and \textsc{Drain} operations, and the terminal conditions leading to either the Shrinker or Amplifier victory. This visualizes the sequential decision-making dynamics and state evolution over time.}
    \label{fig:flux_pipeline}
\end{figure}

These failure modes are consistent with the central hypothesis of the paper: the LLM operates over the observation sequence $\mathbf{x}_{1:t}$, generating the next token $\mathbf{x}_{t+1}$ without maintaining or querying an explicit latent state $\mathbf{s}_t$. The world-model agent, by contrast, computes $\mathbf{s}_{t+1} = f_\theta(\mathbf{s}_t, a_t)$ deterministically and queries value estimates over the resulting state, enabling multi-step planning behavior that is difficult to reliably sustain through pure next-token prediction alone. As illustrated in Figure~\ref{fig:flux_pipeline}, a complete episode of \textsc{Flux} demonstrates the alternating decision process and state evolution under both policies.

\section{Discussion}

\subsection{Empirical Illustration of the $\mathcal{X}$ vs.\ $\mathcal{S}$ Distinction}

The results constitute a case study rather than a general proof. The purpose of the experiment is therefore illustrative rather than comparative in the benchmarking sense. The key result is not that one architecture universally dominates another, but that compiling textual observations into an explicit latent transition system produces qualitatively different planning behavior. Within the controlled setting of FLUX, they illustrate concretely how the same natural-language specification, when compiled into an explicit transition model, yields agents that substantially outperform those reasoning over the observation sequence alone.

The experimental results provide three empirical contributions. First, the RL agent trained exclusively within the extracted world model achieves a $46.2$ percentage-point improvement over random play ($89.5\%$ vs.\ $43.3\%$ win rate), confirming that $\mathcal{S}$-space learning is both feasible and efficient from a formally extracted transition model. Second, the total defeat of the one-step heuristic agent by the RL Amplifier ($100\%$ loss rate) demonstrates that look-ahead depth, enabled by the world model simulator, is the decisive factor separating strategic agents from reactive ones. Third, the consistent $\approx 79\%$ world-model (simulator) win rate against frontier LLMs---across roles, models, and game phases---provides evidence, within this controlled setting, that $\mathcal{X}$-space statistical reasoning alone may be insufficient for reliable dynamic state-tracking and multi-step planning for reliable dynamic state-tracking and multi-step planning in the evaluated setting.

Crucially, both agent classes received identical information: the natural-language rule text. The LLM processed this text as observations; the world-model pipeline used it as a one-time specification to compile an explicit $f_\theta$. Within this environment, the observed performance gap is consistent with the architectural distinction between observation-space sequence modelling and latent-state causal inference---precisely the distinction formalized in the theoretical perspective of this paper.

Importantly, these results should not be interpreted as evidence that LLMs are incapable of understanding or extracting game rules from natural-language context. The successful compilation of the \textsc{Flux} specification into a functional simulator demonstrates that the $ \mathcal{X} \rightarrow \mathcal{S} $ mapping itself is feasible from textual observations. The limiting factor, in this case study, appears to emerge during sequential interaction: the absence of an explicit and persistently updated latent state representation governed by a transition function. Consequently, the primary bottleneck in long-horizon game reasoning is not rule comprehension or the lack of task-specific fine-tuning, but maintaining coherent environment dynamics over extended trajectories.

\subsection{Failure Modes and Latent-State Collapse}

Beyond win rates, the structure of LLM failures carries independent theoretical significance. The $\approx 10\%$ invalid-move rate is not merely a formatting inconvenience; it is consistent with what one might expect from latent-state collapse under the LDI perspective. Each invalid action---an out-of-range index, a malformed operation string, or a reference to a cell removed several turns earlier---reflects a failure to maintain a consistent internal representation of $\mathbf{s}_t$ across the observation sequence. A system with access to an explicit transition model cannot produce such errors by construction, since validity checking reduces to a deterministic causal query over the current state. That LLMs generate them at a non-negligible rate confirms that the model is not querying a maintained world state but instead extrapolating the next token from the distributional pattern of the conversation context.

The declining quality of LLM play over the course of a game reinforces this interpretation. Qualitative inspection of game transcripts reveals a consistent intra-game degradation pattern: early moves are generally legal and locally plausible, whereas moves in the mid-to-late game---precisely where horizon planning and accurate state tracking become decisive---show a marked increase in both invalid outputs and strategically incoherent choices. This trajectory-level deterioration is consistent with the theoretical prediction that $\mathcal{X}$-space models accumulate representational error as the latent trajectory $\tau$ lengthens, since each observation $\mathbf{x}_t$ provides only a partial and potentially ambiguous projection of $\mathbf{s}_t$. The deterministic simulator used in our experiments incurs no such degradation because state transitions are computed exactly: its state estimate is exact at every step by virtue of the deterministic transition $f_\theta$.

The combination of invalid moves, out-of-range references, and strategically incoherent late-game actions suggests progressive divergence between the model's implicit internal trajectory and the true environment state. As conversational context accumulates, the models increasingly behave as though operating on an internally reconstructed approximation of the game rather than the actual board configuration. Invalid outputs and poor strategic decisions therefore appear not as independent phenomena, but as coupled consequences of the same limitation: the absence of a grounded and persistent representation of $\mathbf{s}_t$.

\subsection{Implications for Latent Dynamics Inference}

The perspective developed in this work points toward a fundamental shift in how learning systems should interpret and utilize data. Rather than treating language, vision, or other modalities as independent sources of information, they can be understood as complementary observation channels of a shared latent dynamical system.

\begin{quote}
\textit{
From the perspective of LDI, Intelligence emerges not from modeling observations, but from modeling the latent processes that generate observations.
}
\end{quote}

Under this perspective, observations such as tokens and pixels should be interpreted as indirect evidence about underlying state evolution, rather than primary units of reasoning. Language encodes abstract, high-level trajectories involving causality, intent, and social interaction, while video and other perceptual streams encode low-level physical dynamics such as motion, interaction, and temporal continuity. These modalities provide complementary but incomplete views, each capturing different aspects of the same underlying process.

More broadly, the results are consistent with the central perspective of LDI: that current limitations may arise from an objective-level mismatch in which models optimize for predicting observations rather than recovering the latent processes that generate them. Consequently, sequence models excel at capturing statistical regularities while appearing limited in maintaining causal and dynamical consistency over extended trajectories.

A natural implication of the LDI perspective is the exploration of more state-centric learning approaches that explicitly model latent variables and their transitions across time. Such approaches could treat multimodal data as supervision for learning a shared state space $\mathcal{S}$ governed by coherent transition dynamics. Within this view, learning shifts from surface-level prediction toward inferring structured representations that remain consistent across time, modality, and intervention.

\subsection{Rethinking Corpora and the Design of Learning Systems}

The LDI perspective carries a concrete and actionable implication for the field: the goal of building intelligent systems should not be abandoned sequence modeling, but rather to reorient it. Rather than treating sequence modeling as an end in itself---where the objective is to predict the next observation as accurately as possible---we propose that sequence modeling should be understood as a \textit{means} to an end: a mechanism for extracting latent dynamic knowledge from observational data. This reorientation has two interrelated components.

\paragraph{Corpora as Evidence, Not Data.}
Large text corpora have historically been treated as raw statistical wealth---a source of co-occurrence patterns, distributional regularities, and implicit world knowledge that can be absorbed through scale. Under the LDI perspective, this framing is incomplete. Textual corpora are not merely patterns to be compressed; they are \textit{structured indirect evidence} of underlying dynamical processes: physical interactions, causal chains, agent intentions, belief updates, and social dynamics. Every sentence encodes a partial observation of a latent trajectory in $\mathcal{S}$.

This reframing has a direct consequence for how we evaluate the success of learning systems. A model that achieves high perplexity on held-out text has demonstrated something meaningful, but not the thing that ultimately matters for robust intelligence: whether it has recovered the latent transition structure from which the text was generated. Predictive accuracy over $\mathcal{X}$ is a \textit{proxy} for structural recovery in $\mathcal{S}$---a useful but imperfect one, and one that can be saturated without achieving the underlying goal.

\paragraph{Sequence Modeling as Latent Dynamics Extraction.}
Rather than replacing sequence models, the LDI perspective motivates redesigning them. The central question shifts from ``how accurately can the model predict the next token?'' to ``how faithfully does the model recover the latent state, transition function, and causal structure that generated the observed sequence?'' Under this framing, a language model is not a passive compressor of text but an \textit{active inference engine} whose primary function is to invert the observation process $P(x \mid s)$ and recover the underlying dynamics $P(s \mid x)$.

This has concrete architectural implications. Training objectives that reward latent consistency---such as predicting outcomes of hypothetical interventions, maintaining persistent entity state across long contexts, or generating valid state trajectories under perturbation---would more directly optimize for the latent recovery problem than standard next-token prediction. Similarly, architectures that explicitly disentangle state estimation from observation decoding---mapping inputs into a structured latent representation before generating outputs---are better aligned with the LDI goal than monolithic sequence-to-sequence models.

The three design principles articulated in Section~\ref{sec:ldi}---decoding observations into latent states before reasoning, updating state through transition functions rather than accumulated context, and planning in state space rather than observation space---can be understood as specific instantiations of this general reorientation. They are not merely architectural preferences; they are the structural requirements of any system whose primary objective is to recover and exploit latent dynamics rather than to predict observations.

\paragraph{A Research Imperative.}
The practical implication is that progress in intelligence may depend less on increasing the scale of sequence prediction and more on developing richer forms of supervision and architectural constraints that steer sequence models toward recovering latent dynamics. These might include: (i) multi-step consistency objectives that penalize models for generating sequences inconsistent with inferred state transitions; (ii) grounded evaluation benchmarks that measure latent state recovery rather than surface-level prediction; (iii) hybrid training regimes in which passive language modeling is combined with active interaction in simulated environments to provide direct interventional signal; and (iv) structured latent bottlenecks that force models to compress observations into an explicit and compositionally updatable state representation before generating further outputs.

In sum, the LDI perspective reframes the role of scale and data. The question is not simply how much text we train on, but what machinery we build to extract the dynamical knowledge encoded within it. Language is not the destination---it is the channel. The destination is the latent world model on the other side.

\subsection{Architectural Implications}

From an architectural standpoint, this perspective motivates hybrid systems composed of three tightly coupled components:

\begin{enumerate}
    \item \textbf{Perceptual Encoders (Language and Vision Models):}  
    Modules that map raw observations (tokens, pixels) into structured latent representations, serving as inference mechanisms for hidden states.

    \item \textbf{World Models (Latent Dynamics):}  
    Systems that learn transition functions over latent states, capturing physical laws, causal relationships, and social or cognitive dynamics.

    \item \textbf{Planning and Control Modules:}  
    Decision-making components that operate over the latent state space, enabling simulation, counterfactual reasoning, and long-horizon planning.
\end{enumerate}

Under LDI, this three-component architecture may constitute a useful functional decomposition for systems designed for persistent state estimation, dynamics modeling, and planning, then perceptual encoding, latent dynamics modeling, and planning are not optional modules but the minimal functional decomposition of the inference problem itself.

These components must remain jointly aligned so that observations continuously refine latent state estimates, while predicted transitions are evaluated against new evidence through closed-loop feedback.

Such a synthesis moves beyond next-token prediction toward environment-aware intelligence, where models reason about evolving systems rather than merely describing observed patterns.

\subsection{Limitations}

We acknowledge several limitations of the perspective presented here. First, the analysis assumes a reasonably well-defined latent state space $\mathcal{S}$ and transition function, which may not exist or be tractable for all domains. Second, while we argue that the objective mismatch is structural, the degree to which it matters in practice depends on the specific task and context; LLMs may perform well on many downstream tasks even without explicit world modeling. Third, the \textsc{Flux} environment is intentionally simplified and fully observable, allowing precise isolation of state-tracking and planning behaviour. Although this controlled design strengthens the causal interpretation of the results, more complex partially observable environments will be necessary to evaluate whether the observed gap persists under richer dynamics and broader task distributions.

Most importantly, the results of this study should be interpreted as an illustrative case study of the X vs. S distinction rather than a generalizable empirical claim; a single environment cannot establish that the observed gap holds across task distributions, modalities, or agent architectures.

\section{Conclusion}

This work examined the distinction between observation-space sequence modeling and latent-state world modeling. We argued that many limitations of autoregressive language models—particularly in causal reasoning, long-horizon planning, and dynamic state tracking—stem from an objective-level mismatch: next-token prediction models observations in $\mathcal{X}$ rather than latent environment dynamics in $\mathcal{S}$. To formalize this perspective, we introduced Latent Dynamics Inference, which treats language and multimodal data as indirect observations of underlying state trajectories.

Experiments on \textsc{Flux} offer a proof-of-concept illustration consistent with this perspective. Although both LLMs and RL agents received the same natural-language specification, the world-model agent exhibited substantially more stable behavior in sequential planning tasks within this controlled setting. The observed LLM failure modes---including invalid actions, state-tracking errors, and horizon myopia---are consistent with the absence of a persistently updated latent state, though these results should not be interpreted as a general claim about LLMs versus world models. Crucially, the successful extraction of the \textsc{Flux} simulator from text illustrates that the mapping from observations to latent structure ($\mathcal{X} \rightarrow \mathcal{S}$) is itself feasible from language alone.

Overall, the results suggest that next-token prediction alone may be insufficient for certain forms of long-horizon causal planning unless supplemented with mechanisms for persistent state tracking and transition modeling. More broadly, intelligence may depend less on predicting observations and more on inferring and reasoning over the latent processes that generate them.

We emphasize that LDI is intended as a conceptual perspective rather than a finalized framework or algorithmic prescription. The FLUX experiment serves as a controlled illustrative case study showing how latent transition structure can be extracted from natural-language specifications and used to support explicit state-based planning. Establishing whether the $\mathcal{X}$-space versus $\mathcal{S}$-space distinction generalizes across richer environments, modalities, and architectures remains an open empirical question.

Future work should develop concrete learning objectives and architectural constraints that operationalize the cross-space inference perspective introduced here.

\bibliographystyle{unsrt}  
\bibliography{references}

\end{document}